\documentclass[runningheads]{llncs}
\usepackage{graphicx}
\usepackage{comment}
\usepackage{amsmath,amssymb} 
\usepackage{color}

\usepackage{epsfig}
\usepackage{amsmath}
\usepackage{amssymb}
\usepackage{booktabs}
\usepackage{pifont}
\usepackage{mathrsfs}
\usepackage{dsfont}
\usepackage{wrapfig}
\usepackage[font=small,labelfont=bf,tableposition=top]{caption}

\newcommand{\etal}{\textit{et al.}}

\begin{document}
\pagestyle{headings}
\mainmatter
\def\ECCVSubNumber{2971}  

\title{Learning Trailer Moments in Full-Length Movies with Co-Contrastive Attention} 

\titlerunning{Learning Trailer Moments in Full-Length Movies}
%

\author{Lezi Wang\inst{1}\thanks{This work was done when Lezi Wang worked as an intern at Netflix.} \and
Dong Liu\inst{2} \and
Rohit Puri\inst{3}\thanks{This work was done when Rohit Puri was with Netflix.} \and Dimitris N. Metaxas\inst{1}}
\authorrunning{L. Wang et al.}
%
\institute{$^1$Rutgers University, $^2$Netflix, $^3$Twitch\\
\email{\{lw462,dnm\}@cs.rutgers.edu, dongl@netflix.com, rohipur@twitch.tv}}
\maketitle

\begin{abstract}
A movie's key moments stand out of the screenplay to grab an audience's attention and make movie browsing efficient. But a lack of annotations makes the existing approaches not applicable to movie key moment detection. To get rid of human annotations, we leverage the officially-released trailers as the weak supervision to learn a model that can detect the key moments from full-length movies. We introduce a novel ranking network that utilizes the Co-Attention between movies and trailers as guidance to generate the training pairs, where the moments highly corrected with trailers are expected to be scored higher than the uncorrelated moments. Additionally, we propose a Contrastive Attention module to enhance the feature representations such that the comparative contrast between features of the key and non-key moments are maximized. We construct the first movie-trailer dataset, and the proposed Co-Attention assisted ranking network shows superior performance even over the supervised\footnote{The term ``supervised'' refers to the approach with access to the manual ground-truth annotations for training.} approach. The effectiveness of our Contrastive Attention module is also demonstrated by the performance improvement over the state-of-the-art on the public benchmarks. 
\end{abstract}

\section{Introduction}
\label{sec:intro}
\textit{``Just give me five great moments and I can sell that movie.”} -- Irving Thalberg (Hollywood’s first great movie producer).

Movie is made of moments~\cite{thomson2014moments}, while not all of the moments are equally important. In the spirit of the quote above, some key moments are known as coming attraction or preview, which can not only grab an audience's attention but also convey the movie's theme.

The importance of detecting the key moments is two-fold. First, key moments migrate the content overwhelming. There are millions of movies produced in human history~\cite{huang2018trailers}. A full-length movie typically lasts two or three hours, making it incredibly time-consuming for consumers to go through many of them. The key moments in the form of short video clips can make the movie browsing efficient, where audiences can quickly get the theme by previewing those short clips with story highlightings. Second, for the purpose of movie promotion, the well-selected moments can attract audience to the movie, where the key moments are usually drawn from the most exciting, funny, or otherwise noteworthy parts of the film but in abbreviated form and usually without spoilers\footnote{https://en.wikipedia.org/wiki/Trailer\_(promotion)}. 

A popular form of key moments in the movie industry is the trailer, which is a short preview of the full-length movie and contains the significant shots selected by professionals in the field of cinematography. In this paper, we focus on moments in the movie trailer and try to answer an important question regarding \emph{Movie Trailer Moment Detection} (MTMD) -- can we learn a vision model to detect trailer moments in full-length movies automatically?   

\begin{figure}[t]
  \centering
  \includegraphics[width=0.7\columnwidth]{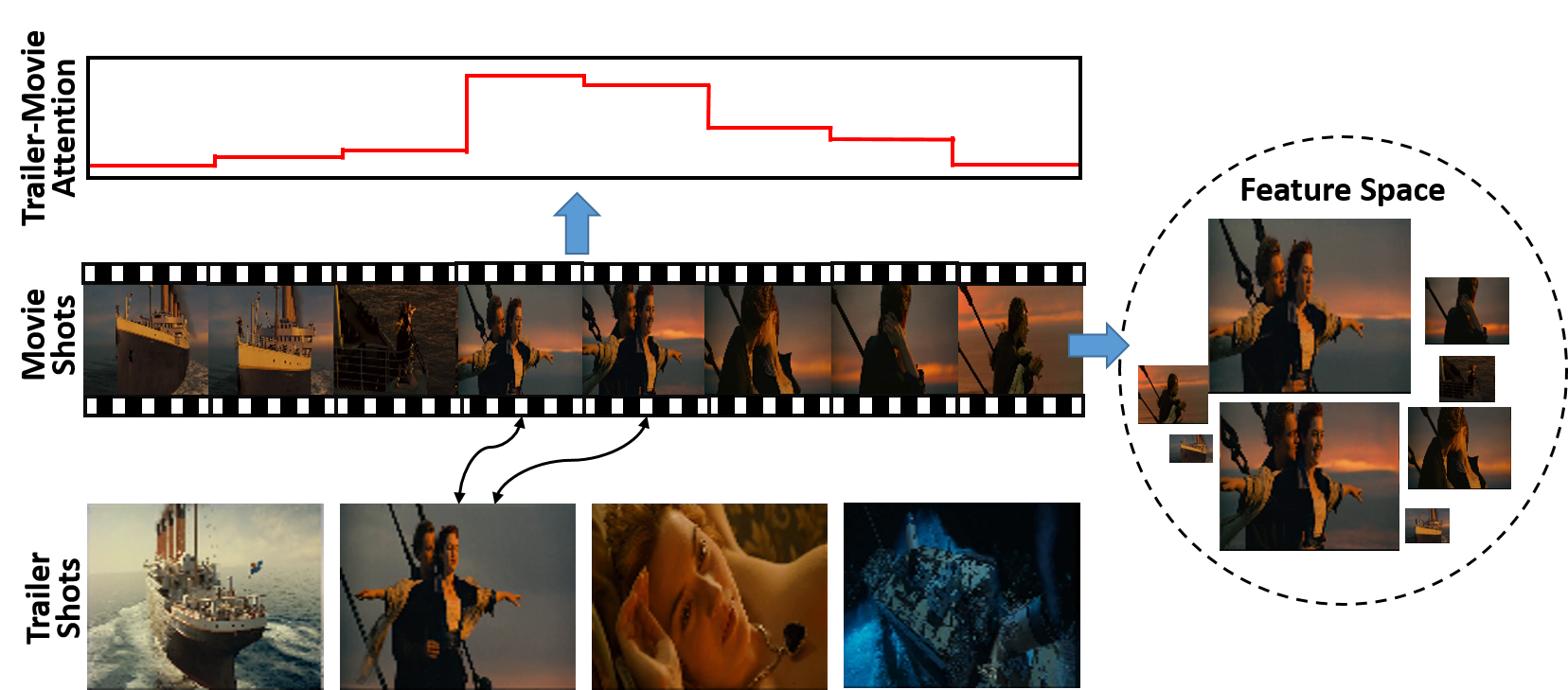}
  \caption{We leverage the trailer shots to estimate the attention scores of individual shots in the full-length movie, which indicate the ``trailerness'' of the shots and can be used as weak supervision to model the contrastive relation between the key and non-key moments in the feature space.}
  \label{fig:teaser}
\end{figure}

The MTMD problem is related to the existing line of research on \emph{Video Highlight Detection} (VHD), a task of extracting highlight clips from videos. Recently, deep learning has become a dominant approach to this task, which formulates it as a problem of learning a ranking model to score the human-labeled highlight clips higher than the non-highlight. Given video clips, the deep spatial-temporal features are extracted as the input to train the ranking model~\cite{carreira2017quo,karpathy2014large,qiu2017learning,wang2018non,xie1712rethinking}. However, the existing VHD approaches cannot be directly applied to MTMD due to the following reasons. 

First, there is no labeled data available for MTMD. To train a robust VHD model, it requires extensive supervision where the annotators must manually identify the highlight clips. Though few efforts have been made to conduct unsupervised VHD, their inferior performance below the supervised indicates the requirement for supervision.  It seems reasonable to annotate the highlights which demonstrate specific actions (e.g., \textit{``Skiing"}, \textit{``Skating"}) or events (e.g., \textit{``Making Sandwich"}, \textit{``Dog Show"}) as in the VHD datasets like \emph{Youtube Highlight}~\cite{sun2014ranking} and TVSum~\cite{song2015tvsum}. However, annotating trailer moments in movies is much more challenging as the selection of trailer moments might attribute to various factors such as emotion, environment, story-line, or visual effects, which requires the annotators to have specialized domain knowledge. To resolve this issue, we create the supervision signal by matching moments between the trailers and the corresponding movies, as shown in Fig.~\ref{fig:teaser}. Specifically, we propose a \emph{Co-Attention} module to measure the coherence between the shots from trailers and movies, through which a set of the best and worst matched shots from the movies are discovered as weakly labeled positive and negative samples.    

Second, the existing VHD approaches treat the individual short clips in the long videos separately without exploring their relations. In fact, the trailer moments follow certain common patterns and should be distinguishable from the non-trailer moments. Taking action movies as an example, although different movies tell different stories, their trailer moments always contain shots with intensive motion activities. To incorporate such prior into MTMD, we propose a \emph{Contrastive Attention module} to enforce the feature representations of the trailer moments to be highly correlated while at the same time encourage the high contrast between the trailer and non-trailer moments. In this way, the features of trailer moments can form a compact clique in the feature space and stand out from the features of the non-trailer moments. 

We integrate the two modules, i.e., \emph{Co-Attention} and \emph{Contrastive Attention}, into the state-of-the-art 3D CNN architecture that can be employed as a feature encoder with a scoring function to produce the ranking score for each shot in the movie. We dub the integrated network \emph{CCANet}: \emph{Co-Contrastive Attention Network}. To support this study and facilitate researches in this direction, we construct TMDD, a Trailer Moment Detection Dataset, which contains $150$ movies and their official trailers. The total length of these videos is over $300$ hours. We conduct experiments on TMDD, and our CCANet shows promising results, even outperforming the supervised approaches. We also demonstrate that our proposed Contrastive Attention module significantly achieves marginal performance-boosting over the state-of-the-art on the public VHD benchmarks, including Youtube Highlight~\cite{sun2014ranking} and TVSum~\cite{song2015tvsum}.

In summary, we make the following contributions:
\begin{itemize}
 \item We propose CCANet that can automatically detect trailer moments from full-length movies without the need of human annotation. 
 \item We propose a Contrastive Attention to constrain the feature representations such that the contrastive relation can be well exploited in the feature space.
 \item To our best knowledge, we are the first to collect a trailer moment detection dataset 
 to facilitate this research direction. 
 \item Our approach shows the superior performance over the state-of-the-art on the public benchmarks,
 outperforming the existing best approach by $13\%$.
\end{itemize}

\section{Related Works}
\textbf{Studies on movie and trailer} have been on the increase interests in computer vision research because of their rich content~\cite{huang2018trailers}. Several efforts have been made to analyze movies or trailers from different angles. A growing line of research is trying to understand the semantics in movies via audio-visual information together with the plot, subtitles, sentiment, and scripts~\cite{bamman2013learning,weng2009rolenet,park2009character,tapaswi2016movieqa,chu2017audio,kang2006space,oosterhuis2016semantic,smith2017harnessing}. The works~\cite{bamman2013learning,weng2009rolenet,park2009character} focus on understanding the relationships of movie characters. Zhu \etal~\cite{zhu2015aligning} proposed an approach to match movie shots and scripts so as to understand high-level storylines. Tapaswi \etal~\cite{tapaswi2016movieqa} developed a movie Q\&A benchmark, proposing a way to understand movies via visual question answering. Chu \etal~\cite{chu2017audio} use machine learning approaches to construct emotional arcs of the visual or audio signals, cluster the type of arcs, and predict audience engagement. Besides the studies on movies, there are also efforts trying to understand the trailers. The works in~\cite{kang2006space,oosterhuis2016semantic} attempt to generate trailers for user-uploaded videos by learning from structures of movies. Smith \etal~\cite{smith2017harnessing} present a heuristic system to fuse the multi-modality to select the candidate shots for trailer creation and the analysis is preformed on horror movies. In~\cite{simoes2016movie,zhou2010movie}, the genre classification problem is investigated by using the trailers to represent the movie content. For this purpose, datasets with several thousand trailers have been constructed. These works are all based on the movie or trailers separately without considering their correspondence. As a pioneering work, Huang \etal~\cite{huang2018trailers} propose an approach to bridge trailers and movies, allowing the knowledge learned from trailers to be transferred to full-length movie analysis. However, a dataset consisting of full-length movies and the key moment annotations is still unavailable, which motivates us to collect TMDD to facilitate this research direction.

\noindent \textbf{Video highlight detection} has been studied a lot for sports videos~\cite{tang2011detecting,xiong2005highlights,rui2000automatically}. Recently, \emph{supervised video highlight detection} has been applied to Internet videos~\cite{sun2014ranking} and first-person videos~\cite{yao2016highlight}. The Video2GIF approach~\cite{gygli2016video2gif} learns to construct a GIF for a video from the user-created GIF-Video pairs. The supervised highlight detection requires human-labeled training pairs, which are expensive to obtain. Recently, several efforts have been made for \emph{unsupervised video highlight detection}, which does not require manual annotations. These approaches can be further divided into \emph{domain-agnostic} or \emph{domains-specific} approaches. The domain-agnostic approaches operate uniformly on any video containing different semantic concepts. The approach in~\cite{mendi2013sports} is based on motion intensity. Works~\cite{panda2017weakly,potapov2014category} are to train a set of video category classifiers and then detect highlights based on the classifier scores or spatial-temporal gradients. In contrast, the domain-specific approaches train highlight detectors on a collection of videos containing the same concept. In \cite{yang2015unsupervised}, Yang \etal propose a category-aware reconstruction loss for unsupervised domain-specific highlight detection.

\begin{wrapfigure}{r}{0.53\columnwidth}
    \centering
    \includegraphics[width=0.5\columnwidth]{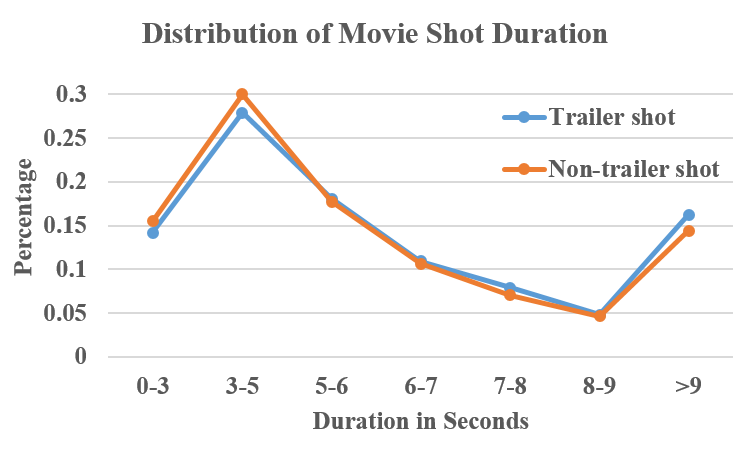}
    \caption{Duration distribution for the trailer and non-trailer shots indicates that the duration of the two kinds of shots is similar.} 
    \label{fig:shot_dur}
\end{wrapfigure}

A very recent work~\cite{xiong2019less} is proposed to get rid of human annotations by leveraging the video duration as the supervision to train highlight detectors. The key insight is that the clips from the shorter user-generated videos are more likely to be the highlights than those from longer videos since users tend to be more focused on the content when capturing shorter videos~\cite{xiong2019less}. While the insight does not apply to movie domain. As shown in Fig.~\ref{fig:shot_dur}, the duration of the trailer and non-trailer shots is similar statistically, which severely mutes the duration signal.

Inspired by the fact that movies come with trailers, we tackle the annotation problem by leveraging the trailer moments to generate the supervision. A Co-Attention module is proposed to measure the coherence between the shots from trailers and movies. Different from the existing Pseudo-Label approach, which offline predicts the labels~\cite{lee2013pseudo,shi2018transductive}, our Co-Attention module is updated in the learning process, where training is in an end-to-end fashion.

\section{Approach}\label{sec:approach}
We develop CCANet with two goals: 1) with the weak-supervision from the publicly available trailers, the network is trained without human labeling; 2) we incorporate the ``contrastive" relation into the learning process so that the trailer moment can be distinguishable from the non-trailer. We first describe how we construct the Trailer Moment Detection Dataset (TMDD) in Sec~\ref{sec:dataset}. Then we present the CCANet in Sec~\ref{sec:ccanet}, consisting of the \textit{Co-Attention} for learning the trailer moments and the \textit{Contrastive Attention} for feature augmentation. 

\subsection{Trailer Moment Detection Dataset}\label{sec:dataset}
We aim to detect the key moments in movies using the publicly available trailers as supervision. However, the existing movie or trailer related benchmarks~\cite{simoes2016movie,zhou2010movie} are not appropriate for this task. They collect the trailers or the movie posters for genre classification without full movies provided. Recently, Huang \etal ~\cite{huang2018trailers} learn the vision models from both movies and trailers by proposing a Large-Scale Movie and Trailer Dataset (LSMTD). However, LSMTD is not publicly available. Moreover, due to the different purposes of learning a semantic model for movie understanding, LSMTD has no ground-truth for MTMD evaluation. To this end, we construct a new dataset, named \emph{Trailer Moment Detection Dataset} (TMDD).

TMDD contains $150$ movies in full length paired with their official trailers. The movies are split into three domains according to the genre, including \textit{``Action"}, \textit{``Drama"}, and \textit{``Sci-Fi"}. Each domain has $50$ movie-trailer pairs. We train an MTMD model for each domain, which accounts for the intuition that the key moments are highly domain-dependent, e.g., a fighting moment might be crucial in \textit{``Action"} movie but not in romantic \textit{``Drama"}. 

We define a movie moment as a shot that consists of consecutive frames in one camera recording time~\cite{smeaton2010video}. We apply the shot boundary detection~\cite{smeaton2010video} to segment movies and trailers into multiple shots. Overall, the TMDD contains $263,837$ movie shots and $15,790$ trailer shots. Hence, MTMD on this dataset is a quite challenging task as the \emph{true positives} only take $\sim6\%$ if we regard all trailer shots as the key moments. 
To our best knowledge, this is the first and largest dataset that has ever been built for MTMD.

To build the ground-truth without the requirement of experts annotating the key moments, we conduct visual similarity matching between trailers and movies at the shot-level and then manually verify the correctness of the matches. The shots occurring both in trailers and full-length movies are regarded as the ground-truth key moments in the movie. Notably, the annotations obtained in this way are only for performance evaluation but not for training the model. In the next section, we present our approach of leveraging the trailers to learn the movie key moments without human annotations needed. 

\subsection{CCANet for Trailer Moment Detection}
\label{sec:ccanet} 
\begin{figure}[t]
  \centering
  \includegraphics[width=0.85\columnwidth]{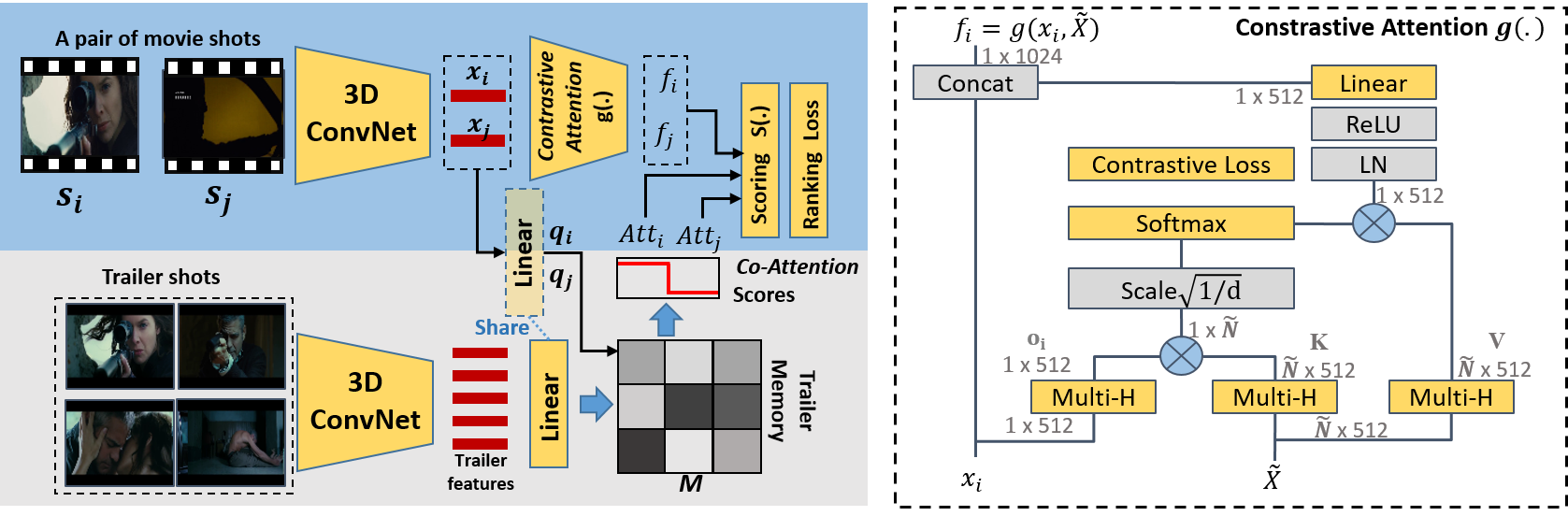}
  \caption{\textbf{Left}: overview of the proposed CCANet. We use the Co-Attention between the trailer and movie as the weak supervision and propose Contrastive Attention to augment the feature representations such that the trailer shots can stand out from the non-trailer shots in the feature space. \textbf{Right:} the details of Contrastive Attention module. $\bigotimes$ indicates matrix multiplication and ``Concat" stands for vector concatenation.}
  \label{fig:framework}
\end{figure}

We integrate the Co-Attention and Contrastive Attention modules into a unified CCANet, as shown in Fig.~\ref{fig:framework}(Left). Our goal is to learn a scoring function $S(\cdot)$ that predicts the ``trailerness'' score of a movie shot given its feature as input, where the feature is extracted from the individual shot by a $3$D ConvNet~\cite{hara2018can}. At test time, movie shots can be ranked based on the predicted scores, and the top-ranked shots are deemed as the key moments that can be applied to create trailers. Specifically, instead of relying on human annotations to create the pairwise shots for learning the $S(\cdot)$, we create shot pairs based on the Co-Attention scores $Att$ between trailers and movies. Additionally, the Contrastive Attention module is proposed to augment the 3D features so as to explore the relations between the trailer and non-trailer shots. The details are descried below.

\subsubsection{Learning Trailer Moments via Co-Attention}\label{sec:framework}

We leverage the Co-Attention between movies and trailers to modify the basic ranking loss for MTMD.\\
\noindent \textbf{Basic Ranking Loss.} We assume that the movie dataset $D$ can be divided into two non-overlapping subsets $D = \{D^+, D^-\}$, where $D^+$ contains the shots of key moments, $D^-$ contains the shots of non-key moment. Let $s_i$ refer to a movie shot and the 3D feature extracted from shot $s_i$
is $x_i$. Since our goal is to rank the shots of key moment higher than the shots of non-key moment, we construct training pairs $(s_i, s_j)$ such that $s_i \in D^+$ and $s_j \in D^-$. We denote the collection of training pairs as $\mathcal{P}$. The learning objective is the ranking loss:
\begin{equation}
\small{
  \mathcal{L}_{Rank} = \sum_{(s_i,s_j) \in \mathcal{P}} \max\big(0, 1 - S(x_i) + S(x_j)\big).}
  \label{eq:rank_sup}
\end{equation}

\noindent \textbf{Co-Attention between Trailer and Movie.} Let $T$ refers to a set of $N_t$ shots in a trailer. We encode each $t_i \in T$ into a 3D feature. As shown in Fig.~\ref{fig:framework}(Left), a linear layer is applied to map the shot features into a memory $M \in \mathbb{R}^{N_t\times d}$, where $d$ is the dimension of the memory vector $m_\tau \in M$. Given the feature $x_i$ of shot $s_i$ from a full movie, we generate the query $q_i$ by applying the linear layer to $x_i$. The Co-Attention can be calculated as the maximal convolution activation between the query $q_i$ and the vectors in $M$:
\begin{equation}
\small{ 
    Att_i = \max_{\tau \in N_t}(q_i \circledast m_{\tau}).
    }
\end{equation}

The Co-Attention score $Att_i$ measures the coherence of shot $s_i$ in the movie to all shots in the trailer $T$. A large $Att_i$ value indicates that the shot $s_i$ is highly correlated to the trailer and therefore is a potential key moment in the movie.

\noindent \textbf{Ranking Loss with Co-Attention.} The ranking loss in Eq.~(\ref{eq:rank_sup}) assumes that we have annotations for constructing the training set $D^+$ and $D^-$. However, it requires extensive human efforts and domain knowledge to annotate them. To achieve the learning goal without access to human annotations, we leverage the trailer to predict the attention score $Att_i$ and use it as a ``soft label'' to measure the importance of shot $s_i$ in the full movie. Additionally, as shown in Fig.~\ref{fig:framework}(Left), we introduce a \emph{Contrastive Attention} module $g(\cdot)$ (described in the next section and illustrated by Fig.~\ref{fig:framework}(Right)) to augment the feature $x_i$ of shot $s_i$ into $f_i$. With the soft labels and augmented features, we can rewrite the learning objective as follows:
\begin{equation}
\small{
\begin{split}
  \mathcal{L}_{Rank} & = \sum_{(s_i,s_j) \in \mathcal{P}} w_{ij} \max \big\{0, 1 - \sigma[S(f_i) - S(f_j)]\big\} \\ 
  & \mathrm{where} \quad w_{ij} = \lambda (\exp(|Att_i - Att_j|) - 1), \\
  & \mathrm{and} \quad \sigma = \mathrm{sgn}(Att_i - Att_j),
\end{split}
}
\label{eq:rank_soft_label}
\end{equation}
where $\lambda$ is a scaling factor and $w_{ij}$ is introduced as a variable to identify the validness of a pair $(s_i, s_j)\in \mathcal{P}$ to the loss. The underlying intuition is that we assign a large weight to the contrastive pair where the difference between $Att_i$ and $Att_j$ is significant and therefore, should be treated as a confident training sample. The variable $\sigma$ is used to determine the order of the predicted scores based on their Co-Attention values. 

It is worth noting that our approach module is different from the existing approach of \emph{learning with Pseudo-Label} (PL). In PL, labels are collected offline from the highly confident predictions made by the model. While our Co-Attention module updates the label predictions in the end-to-end training process.  

\subsubsection{Augmenting Features via Contrastive Attention}\label{sec:CA}
As shown in Fig.~\ref{fig:framework}(Right), we draw inspiration from the attention mechanism~\cite{NIPS2017_7181} to exploit the contrastive relation among shots. Given a target shot $s_i$ and an auxiliary shot set $\tilde{S}$ with $\tilde{N}$ shots, we extract a 3D visual feature $x_i\in \mathbb{R}^{d}$ and a feature set $\tilde{X}\in\mathbb{R}^{\tilde{N}\times d}$, respectively. We apply $\tilde{X}$ as the supportive set to augment $x_i$ to be $f_i = g(x_i,\tilde{X})\in\mathbb{R}^{2d}$. We aim to make the attention contrastive such that the features of key moments can form a compact clique in the feature space and stand out from the features of the non-key moments. Specifically, the attention $A\in\mathbb{R}^{1\times \tilde{N}}$ between $x_i$ and each $x_j\in \tilde{X}$ is computed as:
\begin{equation}
\small{
    A(x_i, \tilde{X}) = \textrm{softmax}(\frac{o_i^TK}{\sqrt{d}}),}
    \label{eq:weight}
\end{equation}
where we use linear layers to map $x_i$ and $\tilde{X}$ to a query vector $o_i$ and key matrix $K$ respectively, and $d$ is the output channel number of the linear layers. The attention score is used to weight the contribution of shots in $\tilde{S}$ to augmenting $s_i$. We apply another linear layer to map $\tilde{X}$ to a value matrix $V$. Then the Contrastive Attention for augmenting $x_i$ to be $f_i$ is formulated as: 
\begin{equation}
\small{
   f_i = \mathrm{concat}\big[x_i,  \textrm{Linear(ReLu}(A(x_i, \tilde{X})\cdot V))}\big].
   \label{eq:fea_aug}
\end{equation}

Now we describe how to construct the auxiliary shot set $\tilde{S}$ for a specific $s_i$ and how to regularize the feature augmentation discussed above. Inspired by our intuition that the cross-video key moments share common patterns and the key and non-key moments in the same video are supposed to be contrastive, we choose both common key moments and non-key moments to construct $\tilde{S}$. In particular, given a shot $s_i$ in a mini-batch during training, we collect all the key moment shot \emph{across videos} as well as the non-key moment shots surrounding $s_i$ in the \emph{same video} into $\tilde{S}$ (More details can be found in the supplementary material). The key and non-key moment shots in the supportive set $\tilde{S}$ are denoted by $\tilde{S}^{+}$ and $\tilde{S}^{-}$ respectively, and we propose the following loss as a regularizer to explicitly impose the contrastive relation between the key and non-key moments:
\begin{equation}
    \mathcal{L}_{C} = -\sum_{i}\theta_i\mathrm{log} \frac{\sum_{j \in \tilde{S}^{+}}\theta_j\exp(o_i^Tk_j)}{\sum_{j \in \tilde{S}^{+}}\theta_j\exp(o_i^Tk_j) + \sum_{j\in \tilde{S}^{-}}(1-\theta_j)\exp(o_i^Tk_j)}
  \label{eq:L_c_att}
\end{equation}

\noindent where $k_j$ is the $j$-th vector in the embedding key matrix $K$ as in Eq.~(\ref{eq:weight}), and $\theta_i$ is a confidence weight indicating the reliability of the soft label for the shot, defined as a function of the Co-Attention score $Att_i$:
\begin{equation}
\small{
 \theta_i = \frac{1}{1 + \exp(-\gamma(Att_i - \epsilon))}
  }
  \label{eq:indicator}
\end{equation}
\noindent where we empirically choose values of $\gamma$ and $\epsilon$ to be $0.65\times \max(Att_i)$ and 100 respectively. Eq.~(\ref{eq:indicator}) approximately maps the Co-Attention score to values of 0 or 1, which is a differentiable function and can be incorporated into the back-propagation of the learning process.

Finally, we combine the Co-Attention ranking loss Eq.~(\ref{eq:rank_soft_label}) and the contrastive loss Eq.~(\ref{eq:L_c_att}) as the training objective of CCANet: 
\begin{equation}
  \mathcal{L} = \mathcal{L}_{Rank} + \mathcal{L}_{C}.
\end{equation}

\section{Experiment Results}\label{sec:results}
\subsection{Movie Key Moment Detection Results}
\noindent \textbf{Dataset and Evaluation Metric.}
We evaluate our CCANet on the constructed dataset TMDD. Under a specific movie genre containing $50$ movies, we randomly split the movies into the training and test set, containing $45$ and $5$ movies respectively. In the experiment, we repeat the split three times and report the average across three runs as the final result. During test, the movie shots are ranked based on the predicted score and then compared with the human-verified ``key moment'' ground-truth obtained by matching shots between trailers and movies as described in Sec.~\ref{sec:dataset}. 

\begin{figure}[t]
  \centering
  \includegraphics[width=0.9\columnwidth]{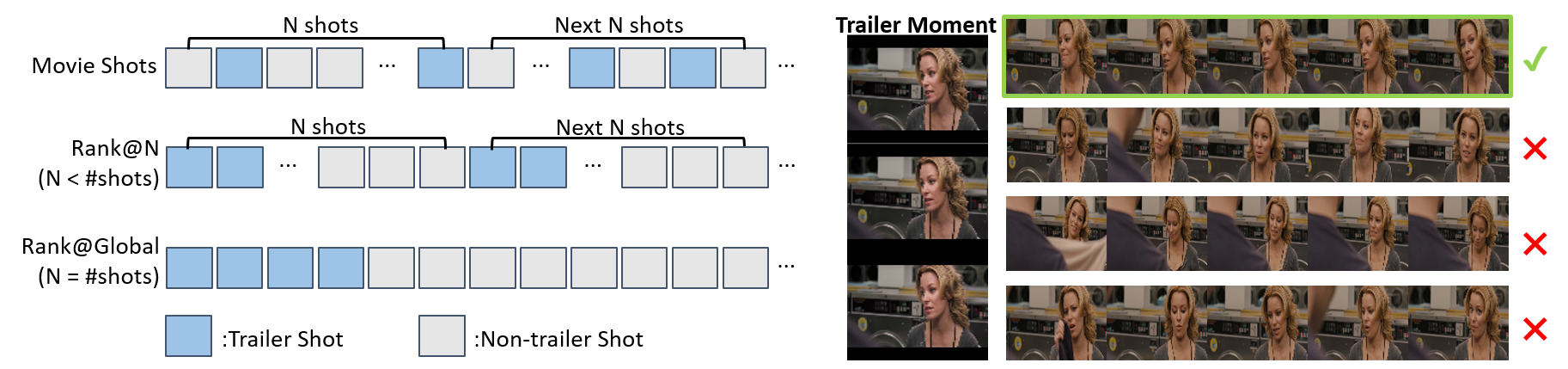}
  \caption{\textbf{Left:} $\mathrm{Rank@N}$. We calculate AP within every $N$ consecutive shots in a full-length movie and average them as the overall performance metric, offering a local-view on the ranking performance. The top row lists trailer (blue) and non-trailer (grey) shots in a movie along the timeline before ranking. The middle and bottom illustrate the ideal $\mathrm{Rank@N}$ results. \textbf{Right:} the ``hard" annotation brings about ambiguity in the labels. A trailer shot and its four visually similar movie shots are shown. The movie shot marked by the green border is labeled as positive and the rest shots are negative.}
  \label{fig:rank_set}
\end{figure}

For the evaluation metric, we calculate Average Precision (AP) on each test video to measure the shot ranking performance. In order to get a fine-grain local view on the ranking performance on each video, we adapt AP to a $\mathrm{Rank@N}$ metric which can be illustrated in Fig.~\ref{fig:rank_set}(Left). As seen, we examine the ranking AP within every $N$ consecutive shots in the movie and average them across the entire movie as the performance metric. $\mathrm{Rank@Global}$ is equivalent to AP where $N$ equals to the number of shots in the movie. We calculate the results on each movie and average them across all test movies as the overall performance.

\noindent \textbf{Feature Extraction.} The $3$D CNN~\cite{hara2018can} (S3D) with a ResNet-$34$~\cite{he2016deep} backbone pre-trained on Kinetics-$400$ dataset~\cite{carreira2017quo} are used to compute the input features. We use the output after the global pooling of the final
convolution layer and a shot is represented by a feature of $512$ dimensions, same as the work~\cite{xiong2019less}. Specifically,  a feature vector is extracted from a snippet covering $16$ consecutive frames. The snippet features are averaged to represent the shot, where a snippet belongs to specific shot if $>70\%$ frames of the snippet are covered by the shot.

\noindent \textbf{Implementation Details.} We implement our model with PyTorch\footnote{https://pytorch.org/}, and optimize the loss with Adam optimizer~\cite{kingma2014adam} for 50 epochs. We use a batch size of $2048$ and set the base learning rate to $0.001$. With a single NVIDIA K80 gpu, the total feature extraction time for a $4$-second shot is $0.05$s. After extracting features, the time to train a ranking model for \textit{Drama} movies is one hour, which contains $480K$ snippets in a total duration of $\sim$$100$ hours. At test time, it takes $0.04$s to score a batch of snippets after feature extraction.

\noindent \textbf{Comparison Baselines.}
We compare our CCANet to two baselines, where the training and inference settings such as learning rate, batch size and so on, follow the same practice as CCANet.
\begin{itemize}
  \item \textbf{Fully Supervised MTMD.}
We assume the annotated trailer shots are accessible. Then we can perform supervised training as the VHD approaches described in Sec~\ref{sec:intro}. The movie shots annotated as trailer moment are the positive samples. For each positive sample, we sample 20 negative (non-trailer) shots, forming a set of pairs to train the ranking model as in Eq.~(\ref{eq:rank_sup}). 
  \item \textbf{Weakly Supervised MTMD with Pseudo Label.}
We also compare CCANet to a weakly supervised approach using the Pseudo Label, which does not require access to manual annotations. We offline calculate the visual similarity between trailer and movie shots. The movie shots having the high similarity to the trailer are regarded as the positive samples, and those with low similarity as the negatives.
\end{itemize}

\begin{table*}[t]
\resizebox{\columnwidth}{!}{
\begin{tabular}{l|cccc|cccc|cccc}
       & Act$_{10}$ & Dra$_{10}$ & ScF$_{10}$ & Avg$_{10}$  & Act$_{20}$ & Dra$_{20}$ & ScF$_{20}$ & Avg$_{20}$ & Act$_{GL}$ & Dra$_{GL}$ & ScF$_{GL}$ & Avg$_{GL}$\\ \hline
Sup    & 0.691         & 0.603        & 0.562         & 0.619          & 0.558         & 0.507        & 0.363         & 0.476          & 0.153         & 0.158        & 0.141         & 0.151          \\ 
Sup+CA & \textbf{0.725}         & 0.641        & 0.589         & 0.652          & 0.583         & 0.524        & 0.382         & 0.496          & 0.171         & 0.163        & 0.153         & 0.162          \\ \hline
PL     & 0.681         & 0.591        & 0.515         & 0.596          & 0.542         & 0.506        & 0.361         & 0.460          & 0.218         & 0.191        & 0.169         & 0.193          \\ 
PL+CA  & 0.714         & 0.625        & 0.548         & 0.629          & 0.577        & 0.539        & 0.372         & 0.486          & 0.269         & 0.215        & \textbf{0.212}         & 0.232          \\ \hline
CoA    & 0.695         & 0.667        & 0.556         & 0.639          & 0.574         & 0.540        & 0.397         & 0.504          & 0.228         & 0.221        & 0.176         & 0.208          \\
CCANet  & 0.723   & \textbf{0.692}        & \textbf{0.591}         & \textbf{0.669}          & \textbf{0.612}         & \textbf{0.562}        & \textbf{0.428}         & \textbf{0.534}          & \textbf{0.271}         & \textbf{0.246}        & 0.210         & \textbf{0.242}       \\
\end{tabular}
}
\caption{The trailer moment detection results on TMDD. ``Sup", ``PL" and ``CoA" denote the different approaches, including fully-supervised, Pseudo-Label and our Co-Attention, with the basic 3D features ~\cite{hara2018can}. The ``Sup+CA", ``PL+CA" and ``CCANet" denote that the shot features in `Sup", ``PL" and ``CoA" are augmented with our proposed Contrastive Attention module. The terms ``Act", ``Dra" and ``ScF" refer to the movie categories, i.e., \emph{Action}, \emph{Drama} and \emph{Sci-Fi}, and ``Avg'' indicates the ``Average'' result across categories. The subscripts ``10", ``20" and ``GL" indicate different evaluation metrics of $\mathrm{Rank@10}$,$\mathrm{Rank@20}$ and $\mathrm{Rank@Global}$.}
\label{tb:res_mtdd}
\end{table*}

\noindent \textbf{Results.} Table~\ref{tb:res_mtdd} presents the trailer moment detection results of different approaches. As seen, by using our Co-Attention (CoA) module alone, our approach substantially outperforms the two baselines. Notably, CoA achieves $\sim$$6\%$ $\mathrm{Rank@Global}$ margin over the supervised approach. The trend is that the $\mathrm{Rank@N}$ drops as $N$ increases, and $\mathrm{Rank@Global}$ is the lowest compared to $N$=$10,20$. The performance drop is attributed to the fact that increasing $N$ involves more negative samples for ranking. Especially, the fully-supervised approach drops the most at the global ranking metric. An explanation is that it suffers from the ``hard" annotations provided by annotators. The ``hard" means that a movie shot is considered as a positive sample only when it is an exact trailer moment. As shown in Fig.~\ref{fig:rank_set} (Right), only the shot at the top is annotated as the trailer shot (positive sample) as it is an exact match to the trailer while the other three are regarded as negative samples. Forcing those movie shots to be separable largely in the feature space brings the ambiguity to train the model. Our CoA module tackles this problem by assigning the soft labels to the data and a training pair with the closer attention scores contributes less to the loss calculation.

We also apply the proposed Contrastive Attention module to augmenting the features in all comparison approaches. In Table~\ref{tb:res_mtdd}, the models with augmented features show superior performance over their origins with the 3D features only~\cite{hara2018can}. The results validate that exploring the relations among different shots can enhance the feature representation and boost the performance.

\begin{figure}[!t]
\begin{minipage}{\textwidth}
  \begin{minipage}[b]{0.49\textwidth}
    \centering
    \includegraphics[width=0.9\columnwidth]{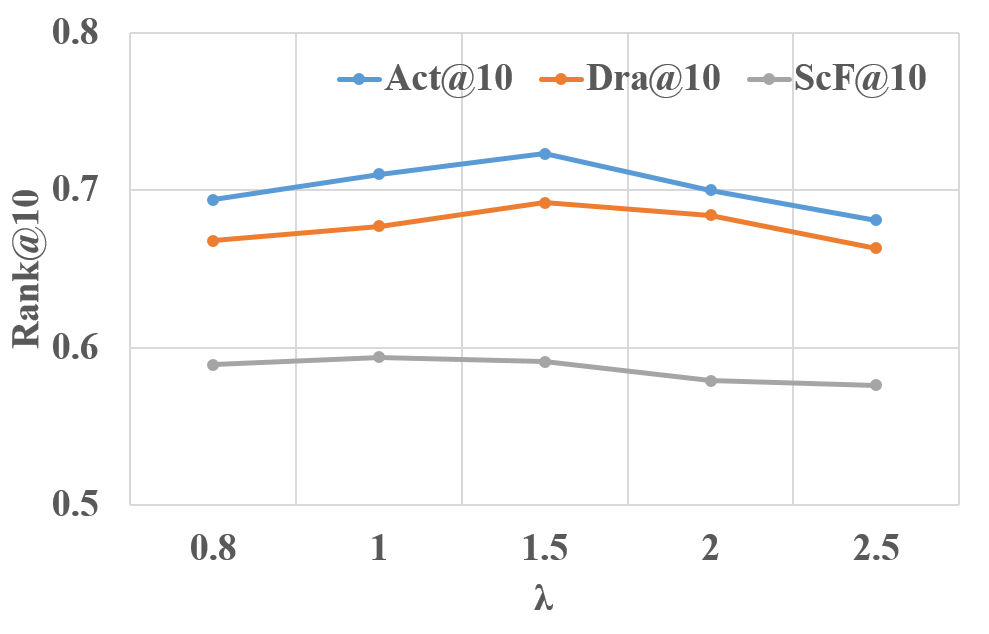}
    \captionof{figure}{Performance variance with respect to $\lambda$. We change the $\lambda$ value and report the performance of the proposed CCANet. The evaluation metric is $\mathrm{Rank@10}$.}
    \label{fig:lambda}
  \end{minipage}
  \hfill
  \begin{minipage}[b]{0.49\textwidth}
    \centering
    \resizebox{\columnwidth}{!}{
    \begin{tabular}{|l|c|c|c|}
    \hline
                        &Act$_{GL}$  &Dra$_{GL}$    & ScF$_{GL}$  \\ \hline
    CoA                 & 0.228 & 0.221  &0.208\\ \hline
    +$FeaAug$          & 0.255 & 0.230  &0.236\\ \hline
    +$FeaAug$+$\mathcal{L}_C$    & \textbf{0.271} & \textbf{0.246}  & \textbf{0.242}\\ \hline
    \end{tabular}}
    \vspace{3em}
    \captionof{table}{$Rank@Global$ of the proposed CoA approach with different feature encoding strategies.}
    \label{tab:Ab}
    \end{minipage}
 \end{minipage}
\end{figure}
\noindent \textbf{Impact of parameter $\lambda$.} In Eq.~(\ref{eq:rank_soft_label}), we introduce a heuristic parameter $\lambda$ to weight the validness of a training pair. The impact of $\lambda$ to CCANet's performance is shown in Fig.~\ref{fig:lambda}, where we report the results measured by $\mathrm{Rank@10}$ and choose the value leading to the best performance. As can be seen, the performance is not sensitive to the value variation of $\lambda$ and we set the value of $\lambda = 1.5$ as default.

\noindent \textbf{Ablation study.} In Table~\ref{tab:Ab}, we perform ablation study to examine our key contribution of Contrastive Attention by evaluating the CoA approach with three variants of shot feature encoding: 1) CoA uses the 3D feature only~\cite{hara2018can}; 2) CoA+$FeaAug$ augments features as Eq~(\ref{eq:fea_aug}) without contrastive loss $\mathcal{L}_C$; 3) CoA+$FeaAug$+$\mathcal{L}_C$ is our CCANet. The $\sim$2\% performance gain from +$FeaAug$ over CoA shows the importance of exploring the relations among clips for feature encoding. Further, our CCANet consistently improves CoA+$FeaAug$. Our interpretation is that the loss $\mathcal{L}_C$ is introduced to guide the attention to be contrastive, encouraging the features of trailer shots to form a compact clique in the feature space and more distinguishable from the features of the non-trailer shots. As a result, it relieves the difficulty of learning the rank model and make CCANet achieve the best performance.

\subsection{Video Highlight Detection Results}
We also evaluate the proposed Contrastive Attention\footnote{Our Co-Attention module is not applicable for the VHD task since there are no video pairs in VHD as the trailer-movie pairs in MTMD.} on the VHD benchmarks, demonstrating its effectiveness.  VHD has a similar goal to MTMD, aiming to detect the highlight moments in video which are supposed to be noticeable among the non-highlight moments, which naturally manifest the contrastive relations. We follow the work~\cite{xiong2019less} to choose two challenging public video highlight detection datasets including YouTube Highlights~\cite{sun2014ranking} and TVSum~\cite{song2015tvsum}. The trained highlight detectors are domain-specific~\cite{xiong2019less}.

\noindent \textbf{Datasets.} YouTube Highlights dataset~\cite{sun2014ranking} contains six domain-specific categories: \textit{surfing, skating, skiing, gymnastics, parkour, and dog}. Each domain consists of $\sim$$100$ videos and the total duration is $\sim$$1430$ minutes. Each video is divided into multiple clips and humans annotate whether a clip contains a specific category. TVSum~\cite{song2015tvsum} is collected from YouTube using $10$ queries and consists of $50$ videos in total from domains such as \textit{changing vehicle tire, grooming an animal, making sandwiches, parade}, etc. (see Table~\ref{tb:tvsum}). We follow the works ~\cite{xiong2019less,panda2017weakly} to average the frame-level scores to obtain the shot-level scores, and then
select the top 50\% shots from each video to build the ground-truth. Finally, the highlights selected by our approach are compared with the ground-truth.

\noindent\textbf{Evaluation Metric and Baselines.} We follow the works in~\cite{xiong2019less,sun2014ranking}, using the mean Average Precision (mAP) and mAP at top-5 to evaluate the highlight detection results on Youtube Highlights~\cite{sun2014ranking} and TVSum~\cite{song2015tvsum}, respectively. We compare with eleven state-of-the-art approaches, which are categorized into unsupervised and supervised approaches. Those previous works' results are reported by the original papers. Specifically, We compare with the unsupervised approaches of RRAE ~\cite{yang2015unsupervised}, MBF~\cite{chu2015video}, CVS~\cite{panda2017collaborative}, SG~\cite{mahasseni2017unsupervised}, DeSumNet(DSN)~\cite{panda2017weakly}, VESD~\cite{cai2018weakly} and LM~\cite{xiong2019less}. In particular, the latest approach LM~\cite{xiong2019less} uses the duration signal as the supervision to train a ranking model and training data contains around $10$M Instagram videos. 

We also include the supervised approaches, e.g. KVS~\cite{potapov2014category}, seqDPP~\cite{gong2014diverse}, SubMod~\cite{gygli2015video}, CLA, GIFs~\cite{gygli2016video2gif} and LSVM~\cite{sun2014ranking}. The latent SVM (LSVM)~\cite{sun2014ranking} has the same supervised ranking loss as ours, but LSVM uses the classic visual features while our features are augmented by the Contrastive Attention module. 

\noindent \textbf{Results on Youtube Highlights.} Table \ref{tb:youtube} shows the results on YouTube Highlights dataset~\cite{sun2014ranking}. All the baseline results are quoted from the original papers. Our approach achieves the best performance and substantially improves those baselines with a large margin. Notably, our approach outperforms the following best performing LM~\cite{xiong2019less} by 12.7\% and CLA by 27.5\%, with relative gains of 23\% and 66\%, where both LM and CLA models are trained on the additional $10$M Instagram videos. We also achieve 15.5\% performance gain over the LSVM. The LSVM~\cite{sun2014ranking} trains a ranking model with domain-specific manually annotated data, but its basic visual feature is limited to capture the feature distribution. Our proposed Contrastive Attention module explicitly models the relations between highlights and non-highlights so that highlight feature can form a compact clique in the feature space and stand out from the features of the non-highlights, leading to a more robust ranking model.
\begin{table}[t]
\centering{
\small{
\begin{tabular}{|l|c|c|c|c|c|c|}
\hline
     & \shortstack{ RRAE~\cite{yang2015unsupervised} } & \shortstack{GIFs~\cite{gygli2016video2gif}} & \shortstack{LSVM~\cite{sun2014ranking}}  &
     \shortstack{CLA~\cite{xiong2019less}}  & 
     \shortstack{LM~\cite{xiong2019less}}      & Ours      \\ \hline
dog    & 0.49 & 0.308 & 0.6      & 0.502 & 0.579     & \textbf{0.633} \\ \hline
gymnastic & 0.35 & 0.335 & 0.41     & 0.217 & 0.417     & \textbf{0.825} \\ \hline
parkour  & 0.5  & 0.54 & 0.61     & 0.309 & \textbf{0.67} & 0.623     \\ \hline
skating  & 0.25 & 0.554 & \textbf{0.62} & 0.505 & 0.578     & 0.529     \\ \hline
skiing  & 0.22 & 0.328 & 0.36     & 0.379 & 0.486     & \textbf{0.745} \\ \hline
surfing  & 0.49 & 0.541 & 0.61     & 0.584 & 0.651     & \textbf{0.793} \\ \hline
Avg.  & 0.383 & 0.464 & 0.536     & 0.416 & 0.564     & \textbf{0.691} \\ \hline
\end{tabular}}
}
\caption{The highlight detection mAP on YouTube
Highlight dataset. Avg. is the average mAP over all the domains. Our approach outperforms all the baselines.}
\label{tb:youtube}
\end{table}

\begin{table}[t]
\centering
\resizebox{\columnwidth}{!}{
\begin{tabular}{|l|c|c|c|c|c|c|c|c|c|c|c|}
\hline
                & MBF   & KVS   & CVS   & SG    & DSN   & VESD & seqDPP & SubMod  & CLA   & LM    & Ours \\
                &~\cite{chu2015video}   &~\cite{potapov2014category}   &~\cite{panda2017collaborative}   &~\cite{mahasseni2017unsupervised}    &~\cite{panda2017weakly}   &~\cite{cai2018weakly} &~\cite{gong2014diverse} &~\cite{gygli2015video}  &~\cite{xiong2019less}   &~\cite{xiong2019less}    &  \\ \hline
Vehicle tire    & 0.295 & 0.353 & 0.328 & 0.423 & -     & -    & - & - & 0.294 & 0.559 & \textbf{0.613}    \\ \hline
Vehicle unstuck & 0.357 & 0.441 & 0.413 & 0.472 & -     & -     & - & - & 0.246 & 0.429 & \textbf{0.546}    \\ \hline
Grooming animal & 0.325 & 0.402 & 0.379 & 0.475 & -     & -     & - & - & 0.590 & 0.612 & \textbf{0.657}    \\ \hline
Making sandwich & 0.412 & 0.417 & 0.398 & 0.489 & -     & -     & - & - & 0.433 & 0.540 & \textbf{0.608}    \\ \hline
Parkour         & 0.318 & 0.382 & 0.354 & 0.456 & -     & -     & - & - & 0.505 & \textbf{0.604} & 0.591    \\ \hline
Parade          & 0.334 & 0.403 & 0.381 & 0.473 & -     & -     & - & - & 0.491 & 0.475 & \textbf{0.701}    \\ \hline
Flash mob       & 0.365 & 0.397 & 0.365 & 0.464 & -     & -     & - & - & 0.430 & 0.432 & \textbf{0.582}    \\ \hline
Beekeeping      & 0.313 & 0.342 & 0.326 & 0.417 & -     & -     & - & - & 0.517 & \textbf{0.663} & 0.647   \\ \hline
Bike tricks     & 0.365 & 0.419 & 0.402 & 0.483 & -     & -     & - & - & 0.578 & 0.691 & \textbf{0.656}    \\ \hline
Dog show        & 0.357 & 0.394 & 0.378 & 0.466 & -     & -     & - & - & 0.382 & 0.626 & \textbf{0.681}    \\ \hline
Average         & 0.345 & 0.398 & 0.372 & 0.462 & 0.424 & 0.423 & 0.447 & 0.461 & 0.447 & 0.563 & \textbf{0.628}    \\ \hline
\end{tabular}
}
\caption{The highlight detection top-5 mAP score on TVSum~\cite{song2015tvsum}. The '-' means that mAP value is not provided in the original paper.}
\label{tb:tvsum}
\end{table}

\noindent \textbf{Results on TVSum.} Table \ref{tb:tvsum} shows the results on TVSum dataset~\cite{song2015tvsum}. Our approach outperforms all the baselines by a noticeable margin. In particular, our results achieve $6.5\%$ mAP higher than the following best performing approach LM~\cite{xiong2019less}. Regarding the supervised approaches, we also outperform SubMod~\cite{gygli2015video} by 16.7\%, where the SubMod~\cite{gygli2015video} proposes an adapted submodular function with structured learning for the highlight detection. 
\subsection{Understanding the Co-Contrastive Attention}

\begin{figure}[t]
    \centering
    \includegraphics[width=0.65\columnwidth]{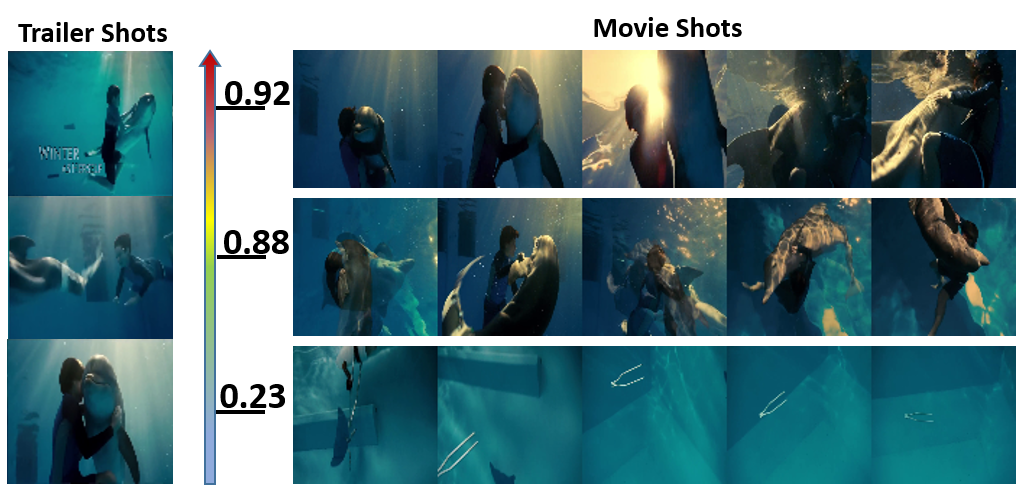}
    \caption{The normalized Co-Attention scores between Trailer and Movie shots. }
    \label{fig:Co_att_example}
\end{figure}

\noindent\textbf{Co-Attention between trailer and movie shots.} We examine the Co-Attention scores between the trailer and movie shots. In Fig.~\ref{fig:Co_att_example}, the score achieves the highest when the trailer moments exactly comes from the movie shots. Our model assigns reasonable high scores to the shots which are visually similar to the trailer moment.

\begin{figure}
    \centering
    \includegraphics[width=\columnwidth]{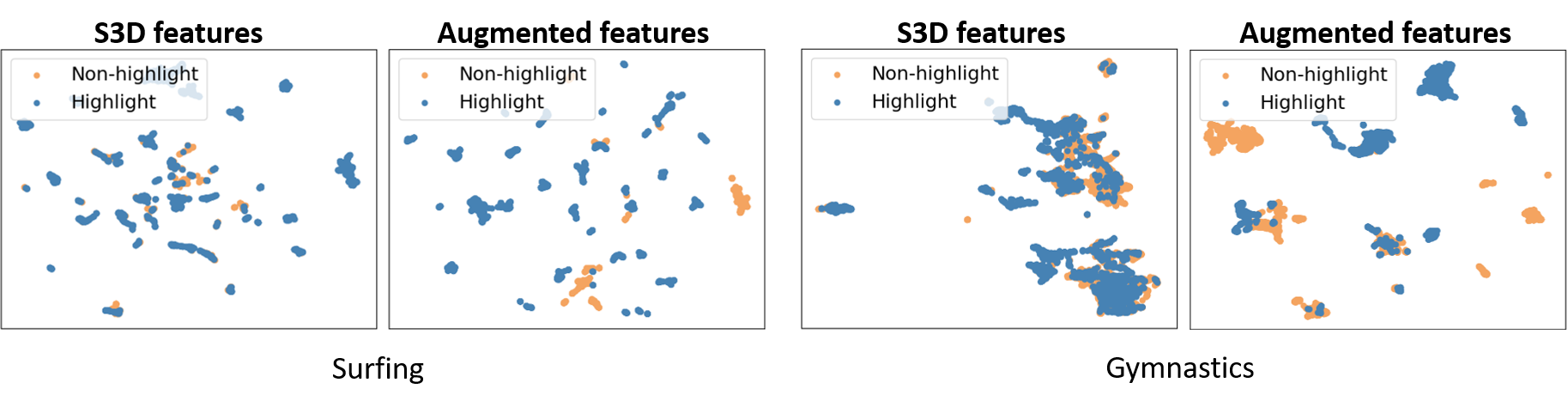}
    \caption{The UMAP visualization of features over domains \emph{``Surfing"} and \emph{``Gymnastics"} (best view in color, zoom in).}
    \label{fig:umap}
\end{figure}

\noindent\textbf{Feature augmented by Contrastive Attention.} In Fig.~\ref{fig:umap}, we plot the UMAP embedding~\cite{mcinnes2018umap} of the basic 3D features and the augmented features with the Contrastive Attention on domains of \emph{``Surfing"} and \emph{``Gymnastic"} from Youtube Highlights~\cite{sun2014ranking} dataset. As can be seen, the augmented highlight and non-highlight features are more separable in the feature space, which eases the difficulty of learning a robust model for highlight detection, resulting in the performance improvement in both domains.

\section{Conclusion}
In this work, we propose the CCANet to address the problem of learning the trailer moments from movies. Our approach utilizes Co-Attention scores as supervision, which does not require expensive human-annotations. Additionally, we introduce the Contrastive Attention module to augment the video features, equipping the model with the capacity of capturing the contrastive relation between the trailer and non-trailer moments. To evaluate our approach, we are the first to collect the dataset, TMDD. The effectiveness of our approach is demonstrated by the performance gain not only on our collected data but also on the public benchmarks. The results on TMDD also demonstrate there is a large room for improvements in trailer moment detection, e.g. multi-modality might be used to boost the robustness, which is part of our future work.  

\clearpage
%
%
\bibliographystyle{splncs04}
\bibliography{egbib}

\begin{thebibliography}{10}
\providecommand{\url}[1]{\texttt{#1}}
\providecommand{\urlprefix}{URL }
\providecommand{\doi}[1]{https://doi.org/#1}

\bibitem{bamman2013learning}
Bamman, D., O’Connor, B., Smith, N.A.: Learning latent personas of film
  characters. In: Proceedings of the 51st Annual Meeting of the Association for
  Computational Linguistics (Volume 1: Long Papers). pp. 352--361 (2013)

\bibitem{cai2018weakly}
Cai, S., Zuo, W., Davis, L.S., Zhang, L.: Weakly-supervised video summarization
  using variational encoder-decoder and web prior. In: Proceedings of the
  European Conference on Computer Vision (ECCV). pp. 184--200 (2018)

\bibitem{carreira2017quo}
Carreira, J., Zisserman, A.: Quo vadis, action recognition? a new model and the
  kinetics dataset. In: proceedings of the IEEE Conference on Computer Vision
  and Pattern Recognition (CVPR). pp. 6299--6308 (2017)

\bibitem{chu2017audio}
Chu, E., Roy, D.: Audio-visual sentiment analysis for learning emotional arcs
  in movies. In: 2017 IEEE International Conference on Data Mining (ICDM). pp.
  829--834. IEEE (2017)

\bibitem{chu2015video}
Chu, W.S., Song, Y., Jaimes, A.: Video co-summarization: Video summarization by
  visual co-occurrence. In: Proceedings of the IEEE Conference on Computer
  Vision and Pattern Recognition (CVPR). pp. 3584--3592 (2015)

\bibitem{gong2014diverse}
Gong, B., Chao, W.L., Grauman, K., Sha, F.: Diverse sequential subset selection
  for supervised video summarization. In: Advances in Neural Information
  Processing Systems (NeurIPs). pp. 2069--2077 (2014)

\bibitem{gygli2015video}
Gygli, M., Grabner, H., Van~Gool, L.: Video summarization by learning
  submodular mixtures of objectives. In: Proceedings of the IEEE Conference on
  Computer Vision and Pattern Recognition (CVPR). pp. 3090--3098 (2015)

\bibitem{gygli2016video2gif}
Gygli, M., Song, Y., Cao, L.: Video2gif: Automatic generation of animated gifs
  from video. In: Proceedings of the IEEE Conference on Computer Vision and
  Pattern Recognition (CVPR). pp. 1001--1009 (2016)

\bibitem{hara2018can}
Hara, K., Kataoka, H., Satoh, Y.: Can spatiotemporal 3d cnns retrace the
  history of 2d cnns and imagenet? In: Proceedings of the IEEE conference on
  Computer Vision and Pattern Recognition (CVPR). pp. 6546--6555 (2018)

\bibitem{he2016deep}
He, K., Zhang, X., Ren, S., Sun, J.: Deep residual learning for image
  recognition. In: Proceedings of the IEEE conference on computer vision and
  pattern recognition (CVPR). pp. 770--778 (2016)

\bibitem{huang2018trailers}
Huang, Q., Xiong, Y., Xiong, Y., Zhang, Y., Lin, D.: From trailers to
  storylines: An efficient way to learn from movies. arXiv preprint
  arXiv:1806.05341  (2018)

\bibitem{kang2006space}
Kang, H.W., Matsushita, Y., Tang, X., Chen, X.Q.: Space-time video montage. In:
  2006 IEEE Computer Society Conference on Computer Vision and Pattern
  Recognition (CVPR). vol.~2, pp. 1331--1338. IEEE (2006)

\bibitem{karpathy2014large}
Karpathy, A., Toderici, G., Shetty, S., Leung, T., Sukthankar, R., Fei-Fei, L.:
  Large-scale video classification with convolutional neural networks. In:
  Proceedings of the IEEE conference on Computer Vision and Pattern Recognition
  (CVPR). pp. 1725--1732 (2014)

\bibitem{kingma2014adam}
Kingma, D.P., Ba, J.: Adam: A method for stochastic optimization. arXiv
  preprint arXiv:1412.6980  (2014)

\bibitem{lee2013pseudo}
Lee, D.H.: Pseudo-label: The simple and efficient semi-supervised learning
  method for deep neural networks. In: Workshop on Challenges in Representation
  Learning, ICML. vol.~3, p.~2 (2013)

\bibitem{mahasseni2017unsupervised}
Mahasseni, B., Lam, M., Todorovic, S.: Unsupervised video summarization with
  adversarial lstm networks. In: Proceedings of the IEEE conference on Computer
  Vision and Pattern Recognition (CVPR). pp. 202--211 (2017)

\bibitem{mcinnes2018umap}
McInnes, L., Healy, J., Melville, J.: Umap: Uniform manifold approximation and
  projection for dimension reduction. arXiv preprint arXiv:1802.03426  (2018)

\bibitem{mendi2013sports}
Mendi, E., Clemente, H.B., Bayrak, C.: Sports video summarization based on
  motion analysis. Computers \& Electrical Engineering  \textbf{39}(3),
  790--796 (2013)

\bibitem{oosterhuis2016semantic}
Oosterhuis, H., Ravi, S., Bendersky, M.: Semantic video trailers. arXiv
  preprint arXiv:1609.01819  (2016)

\bibitem{panda2017weakly}
Panda, R., Das, A., Wu, Z., Ernst, J., Roy-Chowdhury, A.K.: Weakly supervised
  summarization of web videos. In: Proceedings of the IEEE International
  Conference on Computer Vision (ICCV). pp. 3657--3666 (2017)

\bibitem{panda2017collaborative}
Panda, R., Roy-Chowdhury, A.K.: Collaborative summarization of topic-related
  videos. In: Proceedings of the IEEE Conference on Computer Vision and Pattern
  Recognition (CVPR). pp. 7083--7092 (2017)

\bibitem{park2009character}
Park, S.B., Kim, Y.W., Uddin, M.N., Jo, G.S.: Character-net: Character network
  analysis from video. In: Proceedings of the 2009 IEEE/WIC/ACM International
  Joint Conference on Web Intelligence and Intelligent Agent Technology-Volume
  01. pp. 305--308. IEEE Computer Society (2009)

\bibitem{potapov2014category}
Potapov, D., Douze, M., Harchaoui, Z., Schmid, C.: Category-specific video
  summarization. In: European conference on computer vision (ECCV). pp.
  540--555. Springer (2014)

\bibitem{qiu2017learning}
Qiu, Z., Yao, T., Mei, T.: Learning spatio-temporal representation with
  pseudo-3d residual networks. In: proceedings of the IEEE International
  Conference on Computer Vision (ICCV). pp. 5533--5541 (2017)

\bibitem{rui2000automatically}
Rui, Y., Gupta, A., Acero, A.: Automatically extracting highlights for tv
  baseball programs. In: Proceedings of the eighth ACM international conference
  on Multimedia. pp. 105--115. ACM (2000)

\bibitem{shi2018transductive}
Shi, W., Gong, Y., Ding, C., MaXiaoyu~Tao, Z., Zheng, N.: Transductive
  semi-supervised deep learning using min-max features. In: Proceedings of the
  European Conference on Computer Vision (ECCV). pp. 299--315 (2018)

\bibitem{simoes2016movie}
Sim{\~o}es, G.S., Wehrmann, J., Barros, R.C., Ruiz, D.D.: Movie genre
  classification with convolutional neural networks. In: 2016 International
  Joint Conference on Neural Networks (IJCNN). pp. 259--266. IEEE (2016)

\bibitem{smeaton2010video}
Smeaton, A.F., Over, P., Doherty, A.R.: Video shot boundary detection: Seven
  years of trecvid activity. Computer Vision and Image Understanding (CVIU)
  \textbf{114}(4),  411--418 (2010)

\bibitem{smith2017harnessing}
Smith, J.R., Joshi, D., Huet, B., Hsu, W., Cota, J.: Harnessing ai for
  augmenting creativity: Application to movie trailer creation. In: Proceedings
  of the 25th ACM international conference on Multimedia. pp. 1799--1808 (2017)

\bibitem{song2015tvsum}
Song, Y., Vallmitjana, J., Stent, A., Jaimes, A.: Tvsum: Summarizing web videos
  using titles. In: Proceedings of the IEEE conference on computer vision and
  pattern recognition (ICCV). pp. 5179--5187 (2015)

\bibitem{sun2014ranking}
Sun, M., Farhadi, A., Seitz, S.: Ranking domain-specific highlights by
  analyzing edited videos. In: European conference on computer vision (ECCV).
  pp. 787--802. Springer (2014)

\bibitem{tang2011detecting}
Tang, H., Kwatra, V., Sargin, M.E., Gargi, U.: Detecting highlights in sports
  videos: Cricket as a test case. In: 2011 IEEE International Conference on
  Multimedia and Expo (ICME). pp.~1--6. IEEE (2011)

\bibitem{tapaswi2016movieqa}
Tapaswi, M., Zhu, Y., Stiefelhagen, R., Torralba, A., Urtasun, R., Fidler, S.:
  Movieqa: Understanding stories in movies through question-answering. In:
  Proceedings of the IEEE conference on computer vision and pattern recognition
  (ICCV). pp. 4631--4640 (2016)

\bibitem{thomson2014moments}
Thomson, D.: Moments That Made the Movies. Thames \& Hudson (2014),
  \url{https://books.google.com/books?id=\_vNFngEACAAJ}

\bibitem{NIPS2017_7181}
Vaswani, A., Shazeer, N., Parmar, N., Uszkoreit, J., Jones, L., Gomez, A.N.,
  Kaiser, L.u., Polosukhin, I.: Attention is all you need. In: Guyon, I.,
  Luxburg, U.V., Bengio, S., Wallach, H., Fergus, R., Vishwanathan, S.,
  Garnett, R. (eds.) Advances in Neural Information Processing Systems 30
  (NeurIPS), pp. 5998--6008. Curran Associates, Inc. (2017)

\bibitem{wang2018non}
Wang, X., Girshick, R., Gupta, A., He, K.: Non-local neural networks. In:
  Proceedings of the IEEE Conference on Computer Vision and Pattern Recognition
  (CVPR). pp. 7794--7803 (2018)

\bibitem{weng2009rolenet}
Weng, C.Y., Chu, W.T., Wu, J.L.: Rolenet: Movie analysis from the perspective
  of social networks. IEEE Transactions on Multimedia  \textbf{11}(2),
  256--271 (2009)

\bibitem{xie1712rethinking}
Xie, S., Sun, C., Huang, J., Tu, Z., Murphy, K.: Rethinking spatiotemporal
  feature learning for video understanding. Proceedings of the European
  Conference on Computer Vision ECCV  (2018)

\bibitem{xiong2019less}
Xiong, B., Kalantidis, Y., Ghadiyaram, D., Grauman, K.: Less is more: Learning
  highlight detection from video duration. In: Proceedings of the IEEE
  Conference on Computer Vision and Pattern Recognition (CVPR). pp. 1258--1267
  (2019)

\bibitem{xiong2005highlights}
Xiong, Z., Radhakrishnan, R., Divakaran, A., Huang, T.S.: Highlights extraction
  from sports video based on an audio-visual marker detection framework. In:
  2005 IEEE International Conference on Multimedia and Expo (ICME). pp. 4--pp.
  IEEE (2005)

\bibitem{yang2015unsupervised}
Yang, H., Wang, B., Lin, S., Wipf, D., Guo, M., Guo, B.: Unsupervised
  extraction of video highlights via robust recurrent auto-encoders. In:
  Proceedings of the IEEE international conference on computer vision (ICCV).
  pp. 4633--4641 (2015)

\bibitem{yao2016highlight}
Yao, T., Mei, T., Rui, Y.: Highlight detection with pairwise deep ranking for
  first-person video summarization. In: Proceedings of the IEEE conference on
  computer vision and pattern recognition (CVPR). pp. 982--990 (2016)

\bibitem{zhou2010movie}
Zhou, H., Hermans, T., Karandikar, A.V., Rehg, J.M.: Movie genre classification
  via scene categorization. In: Proceedings of the 18th ACM international
  conference on Multimedia. pp. 747--750. ACM (2010)

\bibitem{zhu2015aligning}
Zhu, Y., Kiros, R., Zemel, R., Salakhutdinov, R., Urtasun, R., Torralba, A.,
  Fidler, S.: Aligning books and movies: Towards story-like visual explanations
  by watching movies and reading books. In: Proceedings of the IEEE
  international conference on computer vision (ICCV). pp. 19--27 (2015)

\end{thebibliography}
\end{document}